%% file: iclr2025_delta.tex
\documentclass{article} 
\usepackage{iclr2025_delta,times}

\input{math_commands.tex}

\usepackage{hyperref}
\usepackage{url}
\usepackage{amsmath,amsthm,amssymb}
\usepackage{cleveref}
\usepackage[normalem]{ulem}

\newcommand{\snr}{\mathrm{SNR}} 
\newcommand{\emb}{\Psi}
\newcommand{\femb}{\widehat{\emb}}

\newcommand{\bepsilon}{{\boldsymbol{\epsilon}}}

\def\rvepsilon{{\boldsymbol{\epsilon}}}
\def\rvtheta{{\boldsymbol{\theta}}}
\def\rvphi{{\boldsymbol{\phi}}}

\newcommand{\method}{VQ-LCMD}
\newcommand{\sbest}[1]{\underline{#1}}
\newcommand{\best}[1]{\textbf{#1}}

\usepackage{xcolor}

\usepackage{wrapfig}
\usepackage{multirow,multicol}
\usepackage{microtype}
\usepackage{graphicx}
\usepackage{subfigure}
\usepackage{url}
\usepackage{enumerate}
\usepackage{booktabs}       
\usepackage{stmaryrd}
\usepackage{algorithm}
\usepackage{algorithmic}
\usepackage{minitoc}

\usepackage[toc,page,header]{appendix}

\usepackage{caption}

\hypersetup{colorlinks}

\title{Improving Vector-Quantized Image Modeling with Latent Consistency-Matching Diffusion}

\author{Bac Nguyen$^{1}$\thanks{Correspondence to \texttt{bac.nguyencong@sony.com}}\,\,, Chieh-Hsin Lai$^{1}$, Yuhta Takida$^{1}$, Naoki Murata$^{1}$, Toshimitsu Uesaka$^{1}$, \\\textbf{Stefano Ermon}$^{2}$\textbf{,} \textbf{Yuki Mitsufuji}$^{1, 3}$\\
$^{1}$Sony AI, $^{2}$Stanford University, $^{3}$Sony Group Corporation\\
}

\iclrfinalcopy 
\begin{document}

\doparttoc 
\faketableofcontents 

\maketitle

\begin{abstract}
By embedding discrete representations into a continuous latent space, we can leverage continuous-space latent diffusion models to handle generative modeling of discrete data. However, despite their initial success, most latent diffusion methods rely on fixed pretrained embeddings, limiting the benefits of joint training with the diffusion model. While jointly learning the embedding (via reconstruction loss) and the latent diffusion model (via score matching loss) could enhance performance, end-to-end training risks embedding collapse, degrading generation quality. To mitigate this issue, we introduce \method{}, a continuous-space latent diffusion framework within the embedding space that stabilizes training. \method{} uses a novel training objective combining the joint embedding-diffusion variational lower bound with a consistency-matching (CM) loss, alongside a shifted cosine noise schedule and random dropping strategy. Experiments on several benchmarks show that  the proposed \method{} yields superior results on FFHQ, LSUN Churches, and LSUN Bedrooms compared to discrete-state latent diffusion models. In particular, \method{} achieves an FID of 6.81 for class-conditional image generation on ImageNet with 50 steps.
\end{abstract}
\section{Introduction}

Vector-quantized variational autoencoders (VQ-VAE)~\citep{van2017neural,razavi2019generating} have proven the usefulness of discrete latent representations in image generation~\citep{gu2022vector,chang2022maskgit}. It typically involves training an encoder that compresses the image into a low-dimensional discrete latent space and then using a generative model such as autoregressive models (ARs)~\citep{bengio2000neural,brown2020language} to learn and sample from this discrete latent space. Although ARs appear to dominate discrete data modeling, generating samples from these models incurs significant computational costs. Moreover, controllability is often challenging because the generation order has to be predetermined~\citep{lou2023discrete}, making them less suitable for control tasks such as infilling and inpainting.

On the other hand, continuous-state diffusion models (CSDMs)~\citep{sohl2015deep,ho2020denoising,song2020score} have shown promise as they enable efficient and rapid sampling without relying on the sequential attention mechanism of ARs. Diffusion models learn the inverse of a Markov chain that gradually converts data into pure Gaussian noise, using noise-conditioned score functions (i.e., gradients of log density), which are defined only for continuous data. The core concept is to progressively recover the original data by reversing the diffusion process. Diffusion models are notable for their high-fidelity generation~\citep{dhariwal2021diffusion,lai2023fp,lai2023equivalence}. They offer stable and relatively efficient training procedures that contribute to their success. Recent advances, such as consistency models~\citep{song2023consistency,kim2023consistency,luo2023latent}, have further enhanced diffusion models by reducing the number of sampling steps, making them more practical for real-world applications.

Despite the widespread popularity of CSDMs, their extension to discrete data remains limited. Previous attempts to address this limitation~\citep{austin2021structured,hoogeboom2021argmax,campbell2022continuous,sun2022score,lou2023discrete} have focused on discrete-state diffusion models (DSDMs), which define discrete corruption processes for discrete data and mimic Gaussian kernels used in continuous space. For instance, D3PMs~\citep{austin2021structured} implemented the corruption process as random masking or token swapping and learned to reverse this process from the noisy data. However, unlike continuous diffusion processes, these corruption techniques do not progressively erase the semantic meaning of the data, potentially complicating the learning of the reverse procedure.

Alternatively, discrete data can be mapped into a continuous embedding space~\citep{vahdat2021score,rombach2022high,sinha2021d2c}, followed by the application of CSDMs with typical Gaussian kernels, which enables progressive learning signals~\citep{ho2020denoising} and fine-grained sampling. This approach has been successful in various domains. However, it may not inherently yield satisfactory
results~\citep{li2022diffusion, strudel2022self, dieleman2022continuous}. First, it requires a well-trained embedding for each new discrete dataset~\citep{li2022diffusion} before training CSDMs. Since the embedding space and the denoising model are not trained end-to-end, this can result in suboptimal performance. Second, jointly training both components is challenging and prone to the embedding collapse problem~\citep{dieleman2022continuous,gao2022difformer}, where all embeddings converge to similar vectors.  While this convergence helps the diffusion model predict clean embeddings, it does not result in a meaningful model and instead leads to poor generation. To alleviate embedding collapse, previous work have explored normalizing embedding vectors to a fixed bounded norm~\citep{dieleman2022continuous} or mapping the predicted embedding to its nearest neighbor within the finite set of vectors~\citep{li2022diffusion}. However, the aforementioned manipulations may not yield satisfactory results in practice.

In response, this paper presents \textbf{V}ector-\textbf{Q}uantized  \textbf{L}atent \textbf{C}onsistency-\textbf{M}atching \textbf{D}iffusion (\method{}), a model that enables training of CSDM in discrete vector quantized latent space. We first compress images with a VQ-VAE into discrete tokens and then apply continuous diffusion to their embeddings.  Our key contributions are summarized as follows.
\begin{enumerate}[(i)]
    \item A novel training objective is proposed to stabilize joint training of the embedding and diffusion variational lower bound. In particular, we enforce a \emph{consistency-matching (CM)} loss that requires the model predictions to remain consistent over time. This ensures that the model produces stable outputs throughout the generation process, thereby helping to stabilize training.

    \item We identify several effective techniques to further enhance the generation quality. Specifically, we adopt (1) a shifted cosine noise schedule and (2) random embedding dropout. In addition, we perform a comprehensive analysis to evaluate the empirical impact of these techniques.

    \item Experiments on both unconditional and conditional image generation benchmarks are conducted to evaluate \method{}. The results show that \method{} effectively mitigates the embedding collapse issue and outperforms several baseline methods. \method{} achieves FID scores of $7.25$ on FFHQ, $4.99$ on LSUN Churches, $4.16$ on LSUN Bedrooms, and $6.81$ on ImageNet $256\times256$.
\end{enumerate}

\section{Related Work}\label{sec:literature}
\textbf{Discrete-State Diffusion Models (DSDMs).} The idea is to establish a similar iterative refinement process for discrete data. The corruption process involves transitioning discrete values from one to another. This concept was initially introduced by~\citet{sohl2015deep} for binary sequence problems. Later, it was extended in multinomial diffusion~\citep{hoogeboom2021argmax}. \citet{austin2021structured} improved discrete diffusion by introducing diverse corruption processes, going beyond uniform transition. Based on the former framework, several extensions have been introduced for image modeling, e.g., MaskGIT~\citep{chang2022maskgit}, VQ-Diffusion~\citep{gu2022vector}, Token-Critic~\citep{lezama2022improved}, Muse~\citep{chang2023muse}, and Paella~\citep{rampas2022fast}. Additionally, \citet{campbell2022continuous} utilized Continuous Time Markov Chains for discrete diffusion. Despite their initial success, the corruptions introduced by these methods are characterized by their coarse-grained nature, making them inadequate for effectively modeling the semantic correlations between tokens.

\textbf{Continuous-Space Diffusion Models (CSDMs).} \citet{li2022diffusion} addressed the challenge of controlling language models with Diffusion-LM, a non-autoregressive language model based on continuous diffusion. A similar idea has been introduced in SED~\citep{strudel2022self}, DiNoiSer~\citep{ye2023dinoiser}, CDCD~\citep{dieleman2022continuous}, Bit Diffusion~\citep{Chen2022AnalogBG}, Plaid~\citep{gulrajani2024likelihood}, and Difformer~\citep{gao2022difformer}. The challenge of end-to-end training for both embeddings and CSDMs has not been fully addressed in these methods. To avoid embedding collapse, existing techniques either normalize the embeddings~\citep{dieleman2022continuous} or use heuristic methods~\citep{li2022diffusion}, which are not generally effective and may lead to training instability~\citep{dieleman2022continuous,strudel2022self}.  Recently, \citet{lou2023discrete}  and \citet{meng2022concrete} proposed an alternative concrete score function for discrete settings, which captures the surrogate ``gradient" information within discrete spaces.

\section{Preliminary}
In image synthesis, directly modeling raw pixels can be computationally expensive, especially for high-resolution images.  To reduce this cost, the training process can be divided into two phases. First, an autoencoder is trained to produce lower-dimensional representations, followed by training a generative model in this latent space. This is because pixel-based representations of images contain high-frequency details but little semantic variation~\citep{rombach2022high}. Below, we outline the concept of VQ-VAE and reformulate diffusion models within this discrete latent space.

\subsection{Discrete Representation of Images}
To compress an image into discrete representations, VQ-VAE~\citep{van2017neural} employs a learnable discrete codebook combined with nearest neighbor search to train an encoder-decoder architecture. The nearest neighbor search is performed between the encoder output and the latent embeddings in the codebook. Finally, the resulting discrete latent sequence is then passed to the decoder to reconstruct the image. To further improve the generation fidelity, VQGAN~\citep{esser2021taming} leverages adversarial training to the decoder output.

Given an image, we obtain a sequence of discrete image tokens $\rvx=[\evx_1, \dots, \evx_M]$ with a pre-trained VQ-VAE, where each image token $\evx_i$ belongs to one of the $K$ categories $\{1, \dots, K\}$ in the codebook. Here $M$ denotes the number of image tokens in the discrete space. The distribution over discrete latent variables $\rvx$ is multinomial
and denoted as $P(\rvx)$.

\subsection{Diffusion Models in Discrete Space}
Our goal is to learn a generative model that approximates the probability mass function $P(\rvx)$. To handle discontinuity, we propose using continuous embeddings, where different categories are represented by real-valued vectors.
Let $\rvphi=\{\ve_1, \dots, \ve_K\}$, where $\ve_k \in \mathbb{R}^D$, be a set of vectors, the embeddings of $\rvx$ are then defined as $\emb_\rvphi(\rvx) = [\ve_{\evx_1}, \dots, \ve_{\evx_M}]$. We define a sequence of increasingly noisy versions of $\emb_\rvphi(\rvx)$ as $\rvz_t$, where $t$ ranges from $t=0$ (least noisy) to $t=1$ (most noisy). Next, we adopt the variational diffusion formulation~\citep{kingma2021variational}, incorporating the embedding $\emb_\rvphi$.

\textbf{Forward process.} We define the forward process as a Markov chain, which progressively corrupts the data with Gaussian noise~\citep{kingma2021variational,ho2020denoising}.
For any $t \in [0,1]$, the conditional distribution of $\rvz_t$ given $\rvx$ is modeled as
\begin{align*}
    q_\rvphi(\rvz_t | \rvx) &= \mathcal{N}(\rvz_t | \alpha_t \emb_\rvphi(\rvx), \sigma_t^2 \mI) \,,
\end{align*}
where $\alpha_t$ and $\sigma_t$ are positive scalar-value functions of $t$, which determine how much noise is added to the embeddings of $\rvx$. We consider a variance-preserving process, i.e., $\alpha_t^2 + \sigma_t^2 = 1$. The marginal distribution $q_\rvphi(\rvz_t)$ is a mixture of Gaussian distributions. Due to the Markovian property by construction, the transition probability distributions are given by
\begin{align*}
    q(\rvz_t| \rvz_s) &= \mathcal{N} (\rvz_t | \alpha_{t|s}\rvz_s, \sigma^2_{t|s}\mI) \,,
\end{align*}
where $\alpha_{t|s}=\alpha_t / \alpha_s$ and $\sigma^2_{t|s}=\sigma_t^2 - \alpha^2_{t|s} \sigma_s^2$. Conditioned on the clean discrete variable $\rvx$, the forward process posterior distribution is  derived as
\begin{align*}
    q_{\rvphi}(\rvz_s | \rvz_t, \rvx) = \mathcal{N}(\rvz_s | \mu_{\rvphi}(\rvz_t, \rvx; s, t), \sigma^2(s, t) \mI) \,,
\end{align*}
where $\mu_{\rvphi}(\rvz_t, \rvx; s, t) = \frac{\alpha_{t|s}\sigma^{2}_s}{\sigma^{2}_t}\rvz_{t} + \frac{\alpha_s \sigma^{2}_{t|s}}{\sigma^{2}_{t}}\emb_{\rvphi}(\rvx)$  and $\sigma^{2}(s,t) = \sigma^2_{t|s}\sigma^2_s / \sigma^2_t$.

\textbf{Reverse process.} We gradually denoise the latent variables toward the data distribution by a Markov process. Starting from the standard Gaussian prior $p(\rvz_1)$, the Markov reverse process runs backward from $t=1$ to $t=0$. Let $\rvtheta$ denote the parameters of the denoising model, the conditional probability distribution $p_{\rvphi,\rvtheta}(\rvz_s | \rvz_t; s, t)$ for any $0 \le s \le t \le 1$ in the reverse diffusion process is parameterized by a Gaussian. More specifically, it is given by
\begin{align}
    p_{\rvphi,\rvtheta}(\rvz_s | \rvz_t; s, t) = \mathcal{N}(\rvz_s | \hat{\mu}_{\rvphi,\rvtheta}(\rvz_t; s, t), \sigma^2(s, t) \mI) \,, \label{eq:reparameterization}
\end{align}
where $\hat{\mu}_{\rvphi,\rvtheta}(\rvz_t; s, t) = \frac{\alpha_{t|s}\sigma^{2}_s}{\sigma^{2}_t}\rvz_{t} + \frac{\alpha_s \sigma^{2}_{t|s}}{\sigma^{2}_{t}} \femb_{\rvphi,\rvtheta}(\rvz_t; t)$
and $\femb_{\rvphi,\rvtheta}(\rvz_t; t)$ denotes the predicted embeddings of $\emb_{\rvphi}(\rvx)$ based on its noisy version $\rvz_t$.

\textbf{Network parametrization.} We parameterize $\femb_{\rvphi,\rvtheta}(\rvz_t; t)$ as an average over embeddings, where the $i$-element of $\femb_{\rvphi,\rvtheta}(\rvz_t; t)$ is given by
\begin{align*}
    \big[\femb_{\rvphi,\rvtheta}(\rvz_t; t)\big]_{i} = \sum_{k=1}^{K} P_{\rvtheta}(\tilde{\evx}_i = k| \rvz_t; t) \rve_k\,.
\end{align*}
As the forward process factorizes across $M$ tokens $q_{\rvphi}(\rvz_s | \rvz_t, \rvx) = \prod_{i=1}^M q_{\rvphi}(\rvz_{s,i} | \rvz_{t,i}, \evx_i)$, we also model the reverse process as a factorized distribution. In particular, to estimate the posterior probability $P_{\rvtheta}(\tilde{\rvx}| \rvz_t; t)$, we use  a neural network $f_{\rvtheta}(\rvz_t; t)$ to predict $K$ logits for each token, followed by a softmax nonlinearity, i.e.,
\begin{align*}
   P_{\rvtheta}(\tilde{\rvx}| \rvz_t; t)=\prod_{i=1}^M \mathrm{softmax}([f_{\rvtheta}(\rvz_t; t)]_{i}) \,.
\end{align*}
As $t$ approaches zero, the decoding process from $\rvz_0$ to $\rvx$ gives a learning signal for $\rvphi$.

\textbf{Variational lower bound.} Following~\citep{kingma2021variational}, the negative variational lower bound (VLB) is derived as
\begin{align}
    - \log P_{\rvphi,\rvtheta}(\rvx) &\leq  \E_{\bepsilon}\left[ - \log P_{\rvtheta}(\rvx|\rvz_0; 0) \right] + D_{\text{KL}}(q_{\rvphi}(\rvz_1|\rvx)||p(\rvz_1))
    +
    \mathcal{L}_{\infty}(\rvx; \rvphi, \rvtheta), \label{eq:vlb}
\end{align}
where $\rvz_t = \alpha_t\emb_{\rvphi}(\rvx) + \sigma_t \rvepsilon$ with $\bepsilon \sim \mathcal{N}(\mathbf{0}, \mI)$ and the diffusion loss is simplified to
\begin{align*}
    &\mathcal{L}_{\infty}(\rvx; \rvphi, \rvtheta)=-\frac{1}{2} \mathbb{E}_{\bepsilon,t} \!\!\left[\snr(t)' \|  \emb_{\rvphi}(\rvx) - \femb_{\rvphi,\rvtheta}(\rvz_t; t) \|^2 \right]
\end{align*}
with $\snr(t)=\alpha_t^2/\sigma_t^2$ the signal-to-noise ratio. Under certain conditions\footnote{In theory, we require that $\alpha_1 \emb_\rvphi(\rvx)=0$ to ensure that the prior loss is equal zero.}, the prior loss is close to zero as $q_\rvphi(\rvz_1 | \rvx) \approx \mathcal{N}(0, \mI)$. Unlike CSDMs, the reconstruction loss in our case $\mathcal{L}_{0}(\rvx; \rvphi, \rvtheta)=\E_{\bepsilon}[ - \log P_{\rvtheta}(\rvx|\rvz_0; 0)]$ is important since it involves both denoising and embedding parameters. A remarkable result shown by~\citet{kingma2021variational} is that the diffusion loss is invariant to the noise schedule except at $t = 0$ and $t = 1$.

\begin{figure*}[t]
    \centering
    \includegraphics[width=\textwidth]{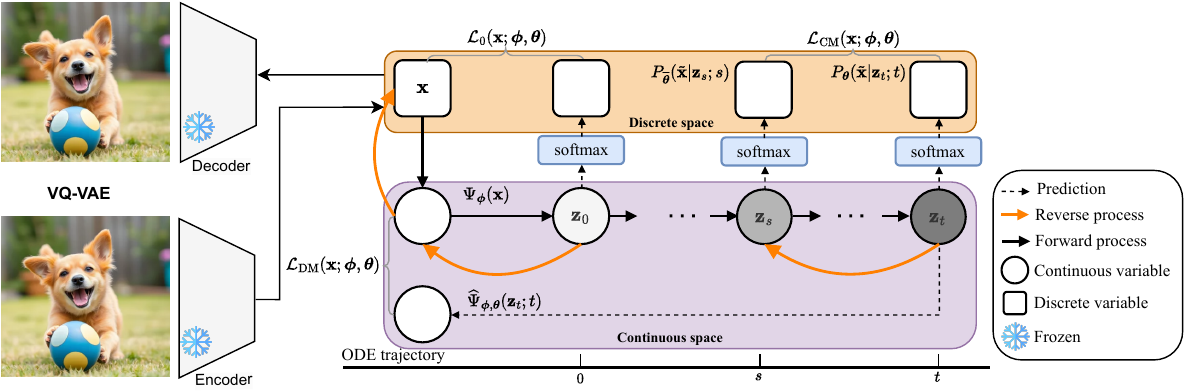}
    \vspace{-15pt}
    \caption{Training procedure of \method{}. An image is compressed into a sequence of discrete tokens $\rvx$ using a pre-trained VQ-VAE. \method{} learns to generate the discrete latent representations $\rvx$ using the consistency-matching (CM) loss, diffusion loss, and reconstruction loss.}
    \label{fig:catdm}
    \vspace{-10pt}
\end{figure*}
Although $\rvphi$ and $\rvtheta$ can be jointly trained by minimizing Eq.~\eqref{eq:vlb}, this approach often results in a solution in which most embeddings collapse into nearly identical vectors with minimal variance, leading to degraded generation quality (refer to our ablation studies in Table~\ref{tab:effect}).

\section{Proposed Method}
This section presents \method{}. First, we introduce the consistency-matching (CM) loss to ensure consistent probability predictions across timesteps. Next, we propose re-weighting the objectives in the loss function, along with an improved noise schedule and a random dropping strategy, to further improve results. 
The overall training and objective function of \method{} is illustrated in Fig.~\ref{fig:catdm}.

\subsection{Consistency-Matching Loss} \label{sec:catdm-cm}
Considering $\emb_\rvphi (\rvx)$ as clean data in the continuous space, the evolution of $\emb_\rvphi (\rvx)$ over time can be described by the probability flow  ordinary differential equation (PF-ODE)~\citep{song2020score}. This PF-ODE allows a deterministic bijection between the embedings $\emb_{\rvphi}(\rvx)$ and latent representations $\rvz_t$. Intuitively, a random noise perturbation $\rvz_t$ of $\emb_\rvphi (\rvx)$ and its relatively nearby point $\overline{\rvz}_s$ along the same trajectory should yield nearly the same prediction.
To ensure these consistent outputs for arbitrary $\rvz_t$, we propose the consistency-matching (CM) loss
\begin{align}\label{eq:consistency_loss}
    &\mathcal{L}_{\mathrm{CM}}(\rvx; \rvphi, \rvtheta) = \mathbb{E}_{\bepsilon, t, s} \left[ D_{\text{KL}} \big(P_{\overline{\rvtheta}}(\tilde\rvx| \overline{\rvz}_s; s) \| P_\rvtheta(\tilde\rvx
 |\rvz_t; t) \big)\right],
\end{align}
where $\overline{\rvtheta}$ denotes an exponential moving average (EMA), i.e., $\overline{\rvtheta} \leftarrow \texttt{stopgrad} (\eta \overline{\rvtheta} + (1 - \eta)\rvtheta )$ with a decay rate of $\eta$. The time variables are sampled uniformly, where $t \sim \mathcal{U}(0,1)$ and $s$ is sampled from the interval $[0,t]$, i.e., $s \sim \mathcal{U}(0,t)$. Here, $\rvz_t$ is obtained by perturbing $\emb_{\rvphi}(\rvx)$ to the noise level $t$ using the transition kernel $q_\rvphi (\rvz_t |\rvx)$ and $\overline{\rvz}_s$ is obtained by taking a PF-ODE step using the EMA model. There are many several ODE solvers to get the PF-ODE step such as Euler~\citep{song2020score} and Heun~\citep{karras2022elucidating} solvers. For simplicity, we use the DDIM sampler~\citep{song2020denoising}, which applies the Euler discretization on the PF-ODE. Under variance preserving settings, it is computed as
\begin{align*}
    \overline{\rvz}_s = \alpha_s\emb_{\overline{\rvphi}}(\rvx) + (\sigma_s/\sigma_t) (\rvz_t - \alpha_t \emb_{\overline{\rvphi}}(\rvx)) \,,
\end{align*}
where $\overline{\rvphi} \leftarrow \texttt{stopgrad} (\eta \overline{\rvphi} + (1 - \eta)\rvphi )$.

Our intuition for the CM loss is that when the timesteps $t$ are small, the model learns the true distribution through the reconstruction loss. As training progresses, this consistency is propagated to later timesteps, eventually reaching $t=1$. The CM loss encourages the probability distributions in neighboring latent variables to converge. Once the model is fully trained, it consistently produces the same probability distribution across the entire PF-ODE trajectory. Since the reconstruction loss enforces the mapping from the embedding space back to discrete data, the learning signal from the reconstruction loss is propagated through the entire PF-ODE trajectory.

\textbf{Connection of CM loss to existing works}. When the distribution $P(\rvx) $ is continuous, Eq.~\eqref{eq:consistency_loss} recovers the consistency training objective in CSDMs~\citep{song2023consistency,kim2023consistency,lai2023equivalence}, which matches clean predictions from models along the same sampling PF-ODE trajectory. Specifically, for any noisy sample $ \rvz_t $ at time $ t $, $ P_{\rvtheta}(\tilde\rvx|\rvz_t; t) $ serves as a deterministic consistency function~\citep{song2023consistency} $\bm{h}_{\rvtheta}(\rvz_t; t) $ predicting the clean sample at time $0$ from $\rvz_t$, regarded as a normal distribution centered around $\bm{h}_{\rvtheta}(\rvz_t; t) $ with small variance. Thus, using the closed-form KL divergence of two normal distributions, Eq.~\eqref{eq:consistency_loss} becomes:
\begin{align*}
\mathcal{L}_{\mathrm{CM}}(\rvx; \rvphi, \rvtheta)\propto \mathbb{E}_{\bepsilon, t, s} \left[ \|\bm{h}_{\overline{\rvtheta}}(\rvz_s; s) - \bm{h}_{\rvtheta}(\rvz_t; t)\|_2^2\right] \,,
\end{align*}
which coincides with a special case of "soft consistency" proposed by~\citet{kim2023consistency} (with their intermediate timesteps $u$ and end time $s$ replaced by our alternate starting time $s$, and our end time $0$). Here, $\propto$ denotes the omission of multiplicative or additive constants that are independent of the training parameters.

\subsection{Final Loss Function}

Although $-\snr(t)'$ in Eq.~\eqref{eq:vlb} provides the correct scaling to treat the objective function as an upper bound of the negative log-likelihood, we hypothesize this weighting function may disrupt the balance between training the reconstruction loss and diffusion loss in practice. Instead of minimizing directly the diffusion loss, we simplify it as
\begin{align*}
\mathcal{L}_{\mathrm{DM}}(\rvx; \rvphi, \rvtheta) = \mathbb{E}_{\bepsilon, t}\left[\| \emb_{\rvphi}(\rvx) - \femb_{\rvphi,\rvtheta}(\rvz_t; t)  \|_2^2 \right]\,.
\end{align*}
This ensures that the loss is evenly distributed over different timesteps. The rationale is that alleviating the error in a large noise level can help the model avoid constant embeddings~\citep{li2022diffusion}.

Putting it all together, the overall objective function of \method{} is given by
\begin{align*}
    \min_{\rvphi, \rvtheta} \quad \E_{\rvx}[\mathcal{L}(\rvx; \rvphi, \rvtheta)] &= \E_{\rvx}[\mathcal{L}_{0}(\rvx; \rvphi, \rvtheta) + \beta_{\mathrm{DM}} \mathcal{L}_{\mathrm{DM}}(\rvx; \rvphi, \rvtheta) +  \beta_{\mathrm{CM}}\mathcal{L}_{\mathrm{CM}}(\rvx; \rvphi, \rvtheta)]\,,
\end{align*}
where $\beta_{\mathrm{DM}} \ge 0$ and $\beta_{\mathrm{CM}} \ge 0$ are hyperparameters. By tuning $\beta_{\mathrm{DM}}$ and $\beta_{\mathrm{CM}}$, we can find the right balance between the objective functions.

\subsection{Noise Schedule}
\begin{wrapfigure}{r}{0.3\textwidth}
    \vskip -0.1in
    \centering
    \includegraphics[width=\linewidth]{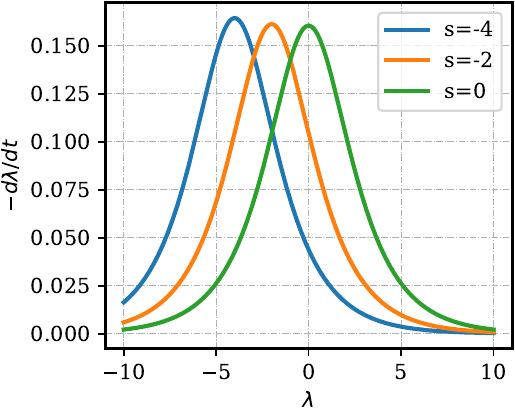}
    \vspace{-15pt}
    \caption{Shifted cosine noise schedule with different shifting factors $s$, $\lambda(t) = \log\snr(t)$.}
    \label{fig:noise_schedule}
\end{wrapfigure}

Although the diffusion loss remains invariant to the noise schedule~\citep{kingma2021variational}, it is essential to determine how noise evolves during the diffusion process~\citep{song2021maximum,kingma2023understanding}. This is because Monte Carlo sampling is employed to estimate the diffusion loss, and thus the training dynamics are influenced by the choice of noise schedule. If the embedding norms are large, denoising would be a trivial task for low noise levels. This is not desired because the denoising model has only a small time window to generate the global structure of the meaningful embedding. To address this, we use the shifted cosine noise schedule~\citep{hoogeboom2023simple}, 
\begin{align*}
    \log\snr(t) = -2\log \tan (\pi t / 2) + s\,,
\end{align*}
where $s \in \mathbb{R}$ is a hyper-parameter. This adjustment changes the noise schedule by shifting its $\log \text{SNR}$ curve. In particular, when $s=0$, it corresponds to the cosine noise schedule~\citep{nichol2021improved}. Essentially, the noise schedule implies different weights in the diffusion loss per noise level~\citep{kingma2023understanding}. As illustrated in Fig.~\ref{fig:noise_schedule}, by moving the curve to the left, it gives more importance for higher degrees of noise.

\subsection{Random Dropping}
Given the noised embeddings, $\mathcal{L}_{\mathrm{CM}}$ used in our training objective ensures the same prediction for the posterior probability $P_\rvtheta(\tilde{\rvx}| \rvz_t; t)$ at any timestep. During joint training, it encourages the model to distinguish the embeddings by increasing their parameter magnitudes. To avoid this shortcut solution, we propose to randomly drop the embeddings. This forces the representations to be more semantic~\citep{he2022masked}. Let $\rvm_\mathrm{RD}\in \{0,1\}^M$ denote a binary mask that indicates which tokens are replaced with a special $\texttt{[mask]}$ token. During training, embeddings of $\rvx$ become $\emb_\rvphi(\rvx\odot\rvm_\mathrm{RD})$. This is because only a portion of the embeddings is used to predict the other tokens. It requires the model to understand the relationship between masked and unmasked tokens. When similar tokens frequently appear in similar contexts, the model learns to associate these tokens closely in the embedding space, as their contextual meanings are similar.

\section{Experiments}
This section evaluates the performance of \method{} on several benchmarks. We begin by outlining the experimental setups, followed by comprehensive experiments covering both conditional and unconditional image generation tasks. Finally, we provide detailed ablation studies to analyze \method{}.

\subsection{Experimental Setup}
We briefly describe the datasets, baselines, and metrics used for evaluation. Additional details are provided in Appendix~\ref{appendix:detail_training}.

\textbf{Datasets.} For unconditional generation, our benchmark consists of three datasets: FFHQ~\citep{karras2019style}, LSUN Bedrooms, and LSUN Churches~\citep{yu2015lsun}.  The FFHQ dataset contains 70K examples of human faces, while the LSUN Bedrooms dataset contains 3M images of bedrooms, and the LSUN Church dataset contains 126K images of churches. For conditional generation, we use ImageNet~\citep{5206848}. These datasets are widely used in the literature. All images have a resolution of $256\times 256$ and VQGAN~\citep{esser2021taming} is used to downsample the images into discrete representations of $16\times 16$ with a codebook size of 1024.

\textbf{Baselines and metrics.} We evaluate \method{} against several baselines, including D3PM with uniform transition probabilities~\citep{austin2021structured}, VQ-Diffusion~\citep{gu2022vector}, and MaskGIT~\citep{chang2022maskgit}. Additionally, we include results for CSDM using fixed embeddings (CSDM$^\dagger$), where embeddings are initialized from the pretrained VQGAN codebook and remain fixed throughout training. For evaluation, we report the Fréchet Inception Distance (FID) between 50,000 generated images and real images. We also provide performance metrics in terms of Precision and Recall~\citep{kynkaanniemi2019improved}. For conditional image generation, we use the Inception Score (IS) as an additional metric to measure the image quality.

\subsection{Unconditional Image Generation} \label{exp:image:generation}

\begin{figure}[t]
    \centering
    \begin{minipage}[t]{0.35\textwidth}
        \vspace{0pt}
        \centering
        \setlength{\tabcolsep}{1.8pt}
        \captionof{table}{Results of ablation studies on the FFHQ dataset}
        \label{tab:effect}
        \centering
        \resizebox{\textwidth}{!}{
            \begin{tabular}{lrrr}
                \toprule
                Method & FID$\downarrow$ & Prec.$\uparrow$ &  Rec.$\uparrow$\\
                \midrule
                CSDM & 77.09 & 0.41 & 0.07 \\
                CSDM w $\ell_2$-norm & 52.35 & 0.55 & 0.12 \\
                \midrule
                \method{} w/o $\mathcal{L}_{\mathrm{CM}}$ & 186.95  & 0.02 & 0.00\\
                \method{} w/o NS & 11.19 & 0.71 & 0.42 \\
                \method{} w/o RD &  \sbest{8.20} & \best{0.73} & 0.42 \\
                \midrule
                \method{} & \best{7.25} & \sbest{0.72} & \best{0.46} \\
                \bottomrule
            \end{tabular}
        }
    \end{minipage}
    \hfill
    \begin{minipage}[t]{0.62\textwidth}
        \vspace{0pt}
        \includegraphics[width=\textwidth]{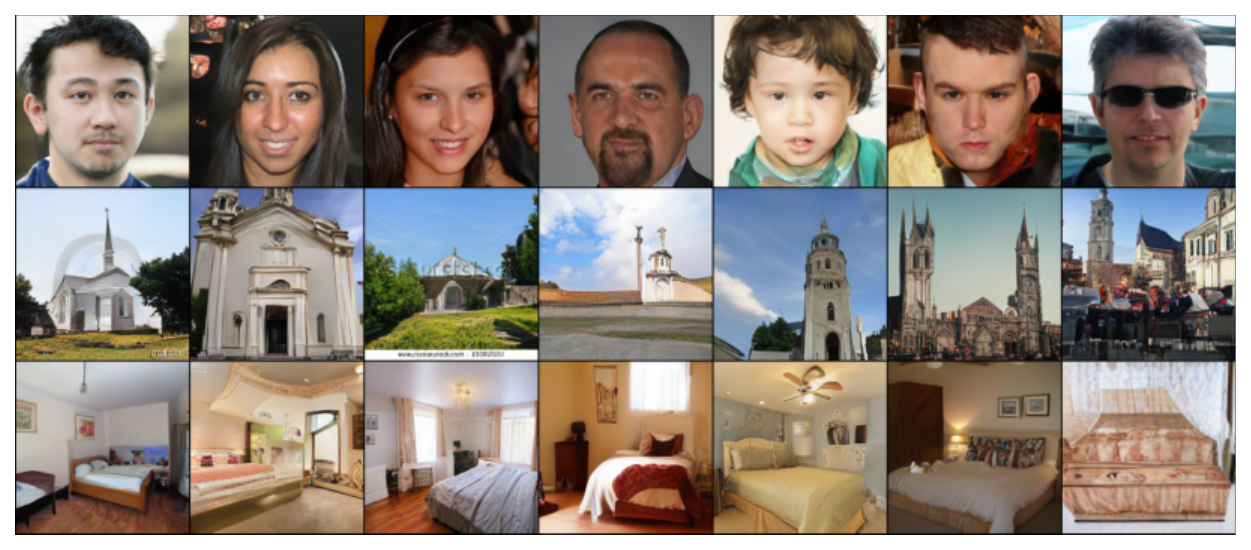}
        \vspace{-20pt}
        \caption{\method{} samples for unconditional generation}
        \label{fig:unconditional}
    \end{minipage}
     \vspace{-15pt}
\end{figure}

Table~\ref{tab:unconditional} presents the results for unconditional image generation tasks. To make a fair comparison, all models are configured with 200 steps for inference. \method{} consistently achieves the lowest FID scores. Furthermore, we investigate the impact of using pretrained embeddings in CSDM$^\dagger$ and demonstrate that while it yields satisfactory results, employing trainable embeddings significantly enhances the performance. On LSUN Bedrooms, \method{} outperforms the baseline methods by a substantial margin, achieving the highest Precision and Recall scores.  These findings underline the superiority of \method{} in generating high-quality samples. The observed improvements in our method compared to discrete diffusion baselines confirm that continuous diffusion models can provide an effective solution for discrete data. Fig.~\ref{fig:unconditional} illustrates samples generated by \method{}.

\begin{table*}[t]
    \small
    \centering
    \caption{Results for unconditional generation on FFHQ, LSUN Churches, and LSUN Bedrooms. The scores of FID, Precision, and Recall are shown. The \best{best} and \sbest{second best} results  are marked.}
    \label{tab:unconditional}
    \begin{tabular}{lccc|ccc|ccc}
    \toprule
    \multirow{2}{*}{Method} & \multicolumn{3}{c|}{FFHQ} & \multicolumn{3}{c|}{LSUN Churches} & \multicolumn{3}{c}{LSUN Bedrooms} \\
    \cmidrule{2-10}
    & FID$\downarrow$ & Prec.$\uparrow$&Rec.$\uparrow$ & FID$\downarrow$ & Prec.$\uparrow$&Rec.$\uparrow$ & FID$\downarrow$ & Prec.$\uparrow$&Rec.$\uparrow$  \\
    \midrule
    \multicolumn{10}{l}{\textbf{\textit{Discrete-Space Diffusion Models}}}  \\
    D3PM Uniform & 9.49 & 0.71 & 0.41 &  6.02 & 0.68 & 0.39 & 6.60 & 0.60 & 0.35\\
    VQ-Diffusion & \sbest{8.79} & 0.70 & \sbest{0.43} & 6.88 & 0.72 & 0.37 & 7.19 & 0.54 & 0.37\\
    MaskGIT & 11.45 & \best{0.75} & 0.42 &  \sbest{5.59} & 0.65 & \best{0.44} & 8.39 & 0.66 & 0.33\\
    \midrule
    \multicolumn{10}{l}{\textbf{\textit{Continuous-Space Diffusion Models}}}  \\
    CSDM$^{\dagger}$ & 12.66 & \sbest{0.73} & 0.38 & 7.88 & \best{0.76} & 0.36 & \sbest{4.93} & \sbest{0.71} & \sbest{0.38} \\
    \method{} (ours) & \best{7.25} & 0.72 & \best{0.46} & \best{4.99} & \sbest{0.75} & \sbest{0.42} & \best{4.16} & \best{0.72} & \best{0.40} \\
    \bottomrule
    \end{tabular}
    \vspace{-10pt}
\end{table*}

\subsection{Conditional Image Generation}
Table~\ref{tab:imagenet} presents the results for class-conditional image generation tasks. To improve the sample quality of conditional diffusion models, we employ the classifier-free guidance~\citep{ho2022classifier}. Essentially, it guides the sampling trajectories toward higher-density data regions. During training, we randomly drop 10\% of the conditions and set the dropped conditions to the null token. Our method achieves a FID of 6.81 and an IS of 225.31 with 50 sampling steps. \method{} notably outperforms both VQGAN and VQVAE-2 by a substantial margin. Compared to MaskGIT\footnote{Since \method{} is implemented in PyTorch, we also use the PyTorch implementation~\citep{besnier2023pytorch} of MaskGIT to ensure a fair comparison.}, \method{} provides competitive FID results and exceeds in IS. However, it is important to note, as highlighted by Besnier and Chen~\citep{besnier2023pytorch}, that MaskGIT requires specific sampling adjustments, such as adding Gumbel noise with a linear decay, to improve its FID. In contrast, \method{} operates without such sampling heuristics.  In addtiion, \method{} performs better than VQ-Diffusion in both FID and IS metrics. For reference samples generated by \method{}, please refer to Appendix~\ref{appendix:samples}. 

\begin{table*}[!thp]
    \footnotesize
    \centering
    \caption{Comparison with generative models on ImageNet $256\times256$. The results of the existing methods are obtained from their respective published works.}
    \label{tab:imagenet}
    \resizebox{\textwidth}{!}{\begin{tabular}{lcccccc}
        \toprule
        Model  & \# params & \# steps & FID$\downarrow$ & IS$\uparrow$ & Precision$\uparrow$ & Recall$\uparrow$ \\
       \midrule
       VQGAN~\citep{esser2021taming} & 1.4B & 256 & 15.78 & 74.3 & n/a& n/a \\
       MaskGIT~\citep{besnier2023pytorch} (PyTorch) & 246M & 8 & \textbf{6.80} & \sbest{214.0} & 0.82 & 0.51 \\
       VQVAE-2\citep{razavi2019generating} & 13.5B & 5120 & 31.11 & 45.00 & 0.36 & \sbest{0.57}\\
       BigGAN-deep~\citep{brock2018large} & 160M & 1 & 6.95 & 198.2 & \textbf{0.87} & 0.28 \\
       Improved DDPM~\citep{nichol2021improved} & 280M & 250 & 12.26 & n/a & 0.70 & \best{0.62}\\
       VQ-Diffusion~\citep{gu2022vector} & 518M &  100 & 11.89 & n/a & n/a & n/a\\
       \midrule
       \method{} (ours) & 246M & 50 & \sbest{6.81} & \best{225.31} & \sbest{0.84} & 0.38 \\
       \bottomrule
    \end{tabular}
    }

\end{table*}

\subsection{Ablation Studies} \label{sec:exp:ablation}
This section presents ablation studies. For additional analysis, please see Appendix~\ref{appendix:additional_ablation}. We investigate the impact of individual components introduced in \method{} on overall performance. Specifically, we examine the shifted cosine noise schedule (NS), random dropping (DR), and consistency-matching loss ($\mathcal{L}_{\mathrm{CM}}$).  The results are presented in Table~\ref{tab:effect}. The baseline method CSDM, trained by minimizing Eq.~\eqref{eq:vlb}, is unable to generate meaningful images. While incorporating an $\ell_2$-norm regularization on the embeddings provides some improvement, it does not completely resolve the collapse issue. \method{} (incorporating our novel components $\mathcal{L}_{\mathrm{CM}}+\text{NS}+\text{RD}$) achieves the best performance. Without RD, the model produces inferior results. Removing NS leads to notable performance degradation. On the other hand, omitting $\mathcal{L}_{\mathrm{CM}}$ results in embedding collapse. These findings highlight the essential role of each component in mitigating the embedding collapse and improving overall performance.

\section{Conclusion}
We have introduced \method{}, a continuous diffusion model tailored for modeling discrete vector-quantized latent distributions, which jointly learns the embeddings and the denoising model. \method{} uses a novel training objective combining the joint embedding-diffusion variational lower bound with a consistency-matching (CM) loss, alongside a shifted cosine noise schedule and random dropping strategy. Experimental results show that \method{} not only alleviates the embedding collapse problem, but also exceeds baseline discrete-state diffusion models.

\textbf{Limitations and future work.} In this work, \method{} is implemented using the Transformer architecture, but we emphasize that the architecture choice is orthogonal to the proposed framework and can be extended to other architectures. Although our main focus is image generation task, \method{} can be applied to any task involving discrete data. Future work will focus on applying it to additional data types, such as graphs and text. It is also interesting to explore more advanced sampling techniques to improve the overall generation quality of \method{}.

\bibliography{iclr2025_delta}
\bibliographystyle{iclr2025_delta}

\newpage
\appendix
\onecolumn
\renewcommand\ptctitle{}

\part{Appendix} 
\parttoc 

\section{Implementation Details} \label{appendix:detail_training}
The prediction network $f_\rvtheta(\rvz_t; t)$ is a bidirectional Transformer~\citep{vaswani2017attention}. For unconditional generation, the network consists of 15 layers, 8 attention heads, and 512 embedding dimensions (a total of 56M parameters). We apply a dropout rate of 0.1 to the self-attention layers. All models are trained on 4 NVIDIA DGX H100 GPUs with a batch size of 128. We use sinusoidal positional embeddings. For conditional generation on ImageNet, we scale up the model to 24 layers, 16 attention heads, and 768 embedding dimensions (a total of 246M parameters). Following~\citep{gu2022vector}, the conditional class label is injected into the model using Adaptive Layer Normalization~\citep{ba2016layer} (AdaLN), i.e., $\text{AdaLN}(\vh, t) = (1 + \va_t) \text{LayerNorm}(\vh) + \vb_t$, where $\vh$ denotes the activation, $\va_t$ and $\vb_t$ are obtained from a linear projection of the class embedding.  We do not use any sampling heuristics such as top-$k$ or nucleus sampling~\citep{Holtzman2020The}.  For random dropping, the dropping probability is fixed to 0.2 as the default. Unless specified otherwise, we set the hyperparameters to $\beta_\mathrm{CM}=1$ and $\beta_\mathrm{DM}=0.005$. For embeddings, we use Gaussian initialization $\mathcal{N}(0, D^{-1/2})$. The EMA rate is set to $\eta=0.99$ and the embedding dimensionality is set to $D=256$.

\section{Latent Variable Classifier-Free Guidance}
It is important to generate images corresponding to a given condition. In \method{}, the condition is incorporated directly into the prediction network through Adaptive Layer Normalization~\citep{ba2016layer}. The assumption here is that the network uses both the corrupted input and the condition to reconstruct the input. However, we often observe that \method{} generates outputs that are not correlated well with the condition. The reason is that the corrupted input contains rich information; therefore, the network can ignore the condition during training.

\begin{figure}[!htp]
    \centering
    \subfigure[$\omega=0$]{\includegraphics[width=0.245\textwidth]{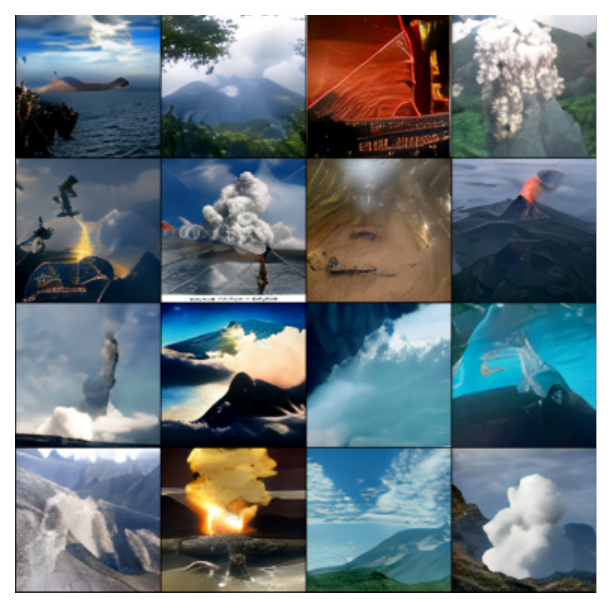}} 
    \subfigure[$\omega=1$]{\includegraphics[width=0.245\textwidth]{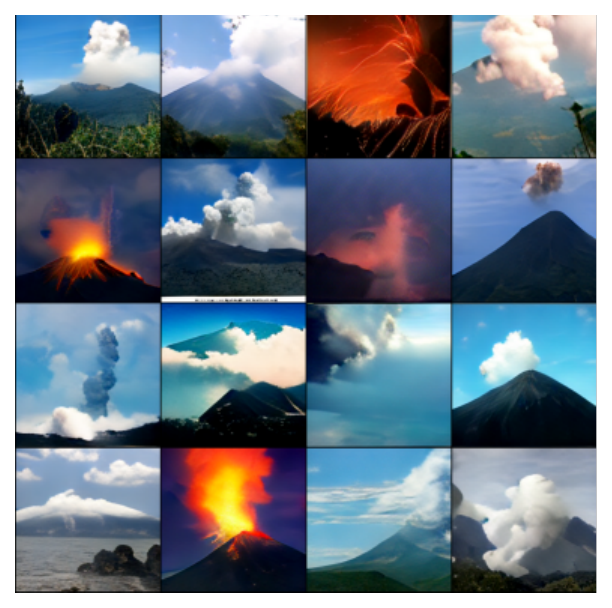}} 
    \subfigure[$\omega=2$]{\includegraphics[width=0.245\textwidth]{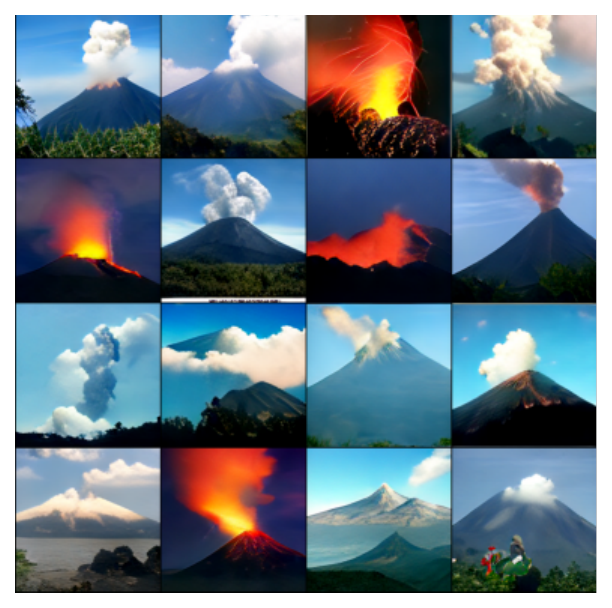}} 
    \subfigure[$\omega=4$]{\includegraphics[width=0.245\textwidth]{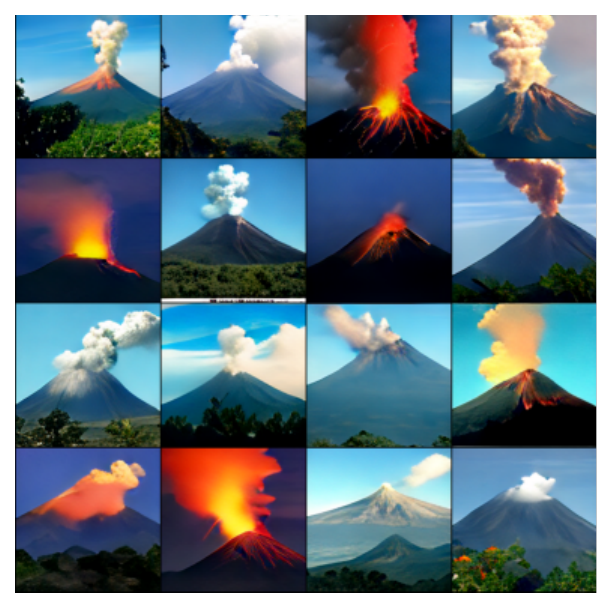}}
    \vspace{-5pt}
    \caption{Generated samples of \method{} with $\omega$ ranging from 0 to 4 on ImageNet.}
    \label{fig:appendix:cfg}
\end{figure}

To improve the sample quality of conditional diffusion models, we employ the classifier-free guidance~\citep{ho2022classifier}. Essentially, it guides the sampling trajectories toward higher-density data regions. During training, we randomly drop 10\% of the conditions and set the dropped conditions to the null token. During sampling, \method{} predicts the categorical variable $\rvx$ as follows
\begin{align}
    \log P_{\rvtheta}(\rvx | \rvz_t, \rvy; t) = (1 + \omega)\log P_{\rvtheta}(\rvx | \rvz_t, \rvy; t) - \omega \log P_{\rvtheta}(\rvx | \rvz_t; t) \,, \label{eq:cfg}
\end{align}
where $\omega \ge 0$ denotes the guidance scale and $\rvy$ denotes the condition. Note that both terms on the right-hand side of Eq.~(\ref{eq:cfg}) are parameterized by the same model. Figure~\ref{fig:appendix:cfg} shows the effects of increasing the classifier-free guidance scale  $\omega$.

\section{Ablation Studies} \label{appendix:additional_ablation}
In this section, we provide additional ablation studies to futher validate our motivations of \method{}.

\subsection{Pretrained vs. Learnable Embeddings}
\begin{figure}[!htbp]
    \centering
     \subfigure[t][Pretrained]{\includegraphics[width=0.23\linewidth]{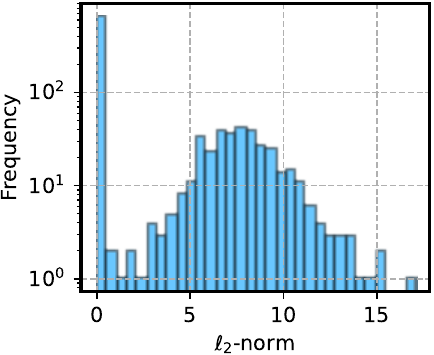}}
     \subfigure[t][Learnable]{\includegraphics[width=0.23\linewidth]{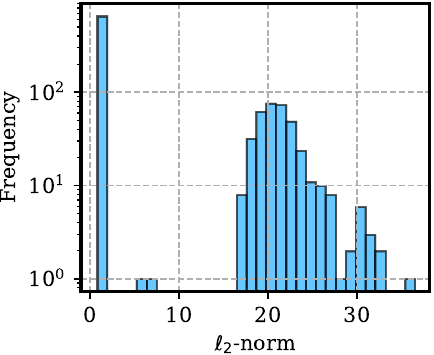}}
     \subfigure[t][Pretrained]{\includegraphics[width=0.245\linewidth]{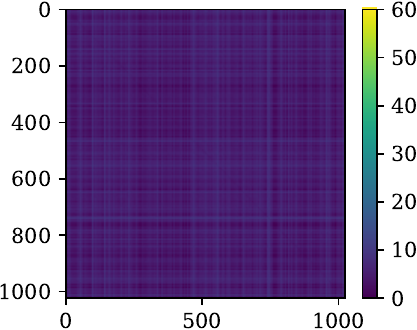}}
     \subfigure[t][Learnable]{\includegraphics[width=0.245\linewidth]{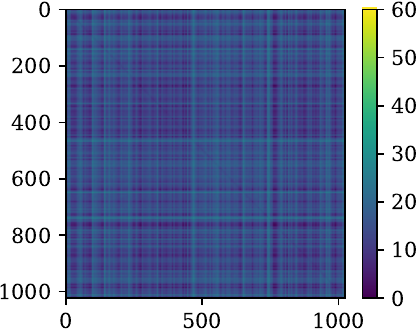}}
     \caption{Visual representation of pretrained and learnable embedding vectors for the LSUN Churches dataset: (a) vector magnitudes for pretrained embeddings, (b) vector magnitudes for learnable embeddings, (c) distance matrix for pretrained embeddings and (d) distance matrix for learnable embeddings. For distance matrices, we compute the Euclidean distances between different embedding vectors.}
     \label{fig:pretrained_vs_learnable}
\end{figure}

We evaluate the embedding vectors obtained by \method{} against those provided by the pretrained VQGAN on the LSUN Churches dataset. Figure~\ref{fig:pretrained_vs_learnable} presents the magnitudes of these vectors and the distance matrices between embeddings. Interestingly, our method learns a structure that is quite similar to the pretrained embeddings. Learnable embeddings tend to have larger magnitudes.

\subsection{Number of Sampling Steps}
We analyze the number of steps necessary to obtain high-fidelity samples. Table~\ref{tab:sampling_steps} presents the FID scores corresponding to different numbers of sampling steps. As expected, we observe a decrease in FID as the number of sampling steps increases. However, the improvement becomes marginal after reaching 50 steps. \method{} can accelerate the conventional diffusion models by a large margin, which is a notable advantage compared to ARs. In addition, we leverage the DDIM sampler to further reduce the number of sampling steps. Table~\ref{tab:maskgit:imagenet} presents a comparison of \method{} using DDIM against MaskGIT~\citep{besnier2023pytorch} (PyTorch implementation). The results show that \method{} achieves a better FID score than MaskGIT when the number of steps is extremely small, highlighting the advantage of our method.

\begin{table}[!htp]
    \centering
    \caption{FID results for different numbers of sampling steps}
    \label{tab:sampling_steps}
    \centering
    \small
    \resizebox{0.7\linewidth}{!}{\begin{tabular}{lccccccc}
        \toprule
         Steps & 5 & 10 & 15 & 20 & 50 & 100 & 200  \\
        \midrule
         Churches & 19.38 & 10.24 & 7.81 & 6.80 & 5.43 & 5.20 & \best{4.99}\\
         Bedrooms & 14.55 & 6.05 & 4.42 & 4.00 & \best{3.86} & 4.01 & 4.16 \\
         FFHQ & 28.80 & 15.55 & 11.44 & 9.57 & 7.56 & 7.34 & \best{7.25}  \\
         \bottomrule
    \end{tabular}}
\end{table}
\begin{table}[!htp]
    \centering
    \caption{FID results on ImageNet for different number of sampling steps}
    \label{tab:maskgit:imagenet}
    \centering
    \resizebox{0.7\linewidth}{!}{\begin{tabular}{lccccccc}
        \toprule
        Steps & 2 & 3 & 4 & 5 &6 &7 & 8  \\
        \midrule
        MaskGIT & 97.83 & 46.43 & 20.29 & 10.95 & 7.74 & \textbf{6.79} & \textbf{6.80}  \\
         \method{} & \textbf{17.83} & \textbf{10.57} & \textbf{8.51} & \textbf{7.87} & \textbf{7.56} & 7.45 & 7.44 \\
         \bottomrule
    \end{tabular}
    }
\end{table}


\subsection{Dropping Strategies}
We explore three different strategies to drop tokens during training. One strategy involves linearly increasing the dropping ratio concerning the timestep (\textbf{linear}). In this scheme, early timesteps involve a small portion of tokens being dropped, while in later timesteps a higher proportion of tokens are dropped. Another strategy is to randomly select a ratio and drop the tokens according to this ratio (\textbf{rand\_drop}).  Finally, a fixed dropping ratio $0\le r \le 1$ can be employed (\textbf{rand(r)}). Table~\ref{tab:masking} summarizes the results. \method{} performs the best with an appropriately chosen fixed dropping ratio.

\begin{table}[!htp]
    \centering
    \caption{Ablation results on different dropping strategies}
    \label{tab:masking}
    \centering
    \resizebox{0.45\textwidth}{!}{\begin{tabular}{lccc}
        \toprule
        & FID$\downarrow$ & Precision$\uparrow$ &  Recall$\uparrow$\\
        \midrule
        linear & 9.12 & \textbf{0.72} & 0.41 \\
        rand\_drop & 8.44 & 0.71 & 0.42 \\
        \midrule
        rand (0.1) &  7.81 & 0.71 & 0.43  \\
        rand (0.2) &  \textbf{7.25} & \textbf{0.72} & \textbf{0.46} \\
        rand (0.3) &  8.45 & 0.70 & 0.43  \\
        rand (0.4) &  9.89 & 0.70 & 0.41  \\
        rand (0.5) &  9.11 & \textbf{0.72} & 0.41  \\
         \bottomrule
    \end{tabular}
}
\end{table}

\subsection{Weighting Terms}
We hypothesize that balancing the reconstruction loss and the diffusion loss is crucial to preventing embedding collapse. In \method{}, this is achieved by tuning the hyperparameter $\beta_{\mathrm{DM}}$. Table~\ref{tab:ablation:weighting} presents the FID results on FFHQ for various combinations of $\beta_{\mathrm{CM}}$ and $\beta_{\mathrm{DM}}$. Adjusting these parameters alters the contributions of the diffusion loss and the consistency-matching loss in the objective function. As indicated in the table, when $\beta_{\mathrm{DM}}$ is relatively large, the model still suffers from embedding collapse.

\subsection{Embedding dimensionality}
Table~\ref{tab:ablation:dimension} shows the influence of embedding dimensionality. We report the FID results on FFHQ when varying the embedding dimensionality. \method{} demonstrates consistent performance across various dimensionalities. As the dimensionality increases, the performance slightly decreases. \method{} achieves the best result when $D=128$.

\begin{table}[!thp]
    \centering
    \begin{minipage}{.45\linewidth}
        \centering
        \caption{Results on $\beta_{\mathrm{CM}}$ and $\beta_{\mathrm{DM}}$}
        \label{tab:ablation:weighting}
        \begin{tabular}{llr}
            \toprule
            $\beta_{\mathrm{CM}}$ &  $\beta_{\mathrm{DM}}$ & FID $\downarrow$\\
           \midrule
           0.01 & 0.01 & 175.46\\
           0.01 & 1 & 173.28 \\
           1 & 1 & 54.10 \\
           1 & 0.01 & 8.26 \\
           1 & 0.005 & 7.25 \\
           \bottomrule
        \end{tabular}

    \end{minipage}
    \hfill
    \begin{minipage}{.45\linewidth}
        \caption{Embedding dimensionality}
        \label{tab:ablation:dimension}
        \centering
        \begin{tabular}{lr}
            \toprule
            $D$ & FID $\downarrow$\\
           \midrule
           64 &  7.90\\
           128 &  7.20 \\
           256 &  7.25 \\
           768 & 7.42 \\
           1024 & 7.38 \\
           \bottomrule
        \end{tabular}

    \end{minipage}
    \hfill
\end{table}

\subsection{Ablation Studies on ImageNet}
\begin{figure}[!htp]
    \centering
     \hfill\subfigure[Varying weights]{\includegraphics[width=0.48\linewidth]{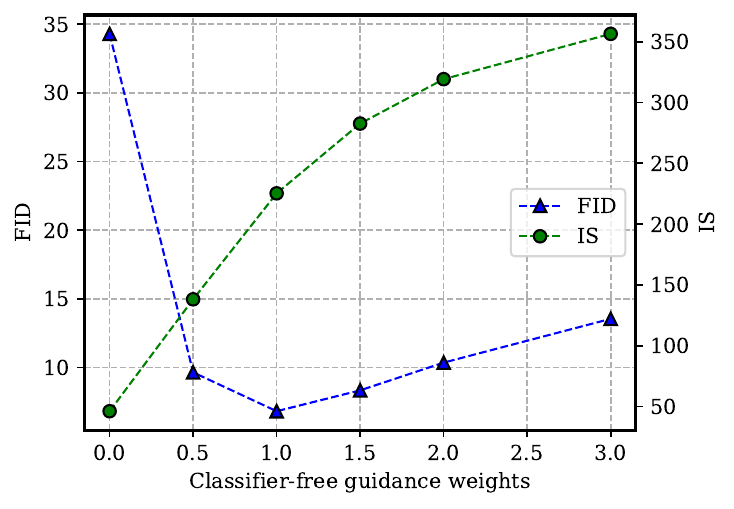}\label{fig:fid_vs_is}}\hfill
     \subfigure[Varying number of sampling steps]{\includegraphics[width=0.48\linewidth]{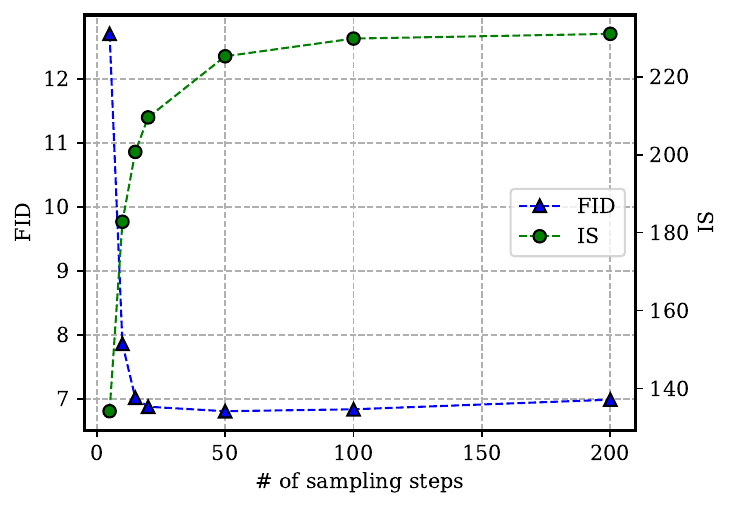}\label{fig:fid_vs_is_step}}\hfill
    \caption{Ablation studies for FID vs IS on ImageNet when (a) varying classifier-free guidance weights and (b) varying number of sampling steps.}

\end{figure}

We conduct ablation studies on ImageNet to examine the effects of classifier-free guidance weights and the number of sampling steps. Figure~\ref{fig:fid_vs_is} shows the FID and IS metrics across various classifier-free guidance weight values. Additionally, Figure~\ref{fig:fid_vs_is_step} presents the FID and IS results as we vary the number of sampling steps. There is a clear trade-off between fidelity represented by FID and quality represented by IS. \method{} achieves the best FID results when $\omega=1$.

\section{Pseudocode}
\begin{figure*}[t]
\begin{minipage}[t]{0.47\textwidth}
\begin{algorithm}[H]
  \caption{Training} \label{alg:training}
  \small
  \begin{algorithmic}[1]
    \REPEAT
      \STATE Sample batch of $\rvx \sim P(\rvx)$
      \STATE $t \sim \mathcal{U}(0, 1)$; $s \sim \mathcal{U}(0, t)$; $\epsilon\sim\mathcal{N}(\mathbf{0},\mI)$; $\rvm_{\text{RD}}\sim \{0,1\}^M$
      \STATE $\rvz_t := \alpha_t (\emb_\rvphi(\rvx) \odot \rvm_{\text{RD}})+ \sigma_t\bepsilon$
      \STATE $\overline{\rvz}_s :=  \alpha_s\emb_{\overline{\rvphi}}(\rvx) + (\sigma_s/\sigma_t) (\rvz_t - \alpha_t \emb_{\overline{\rvphi}}(\rvx))$
      \STATE Take gradient descent step on
      \STATE $\quad \nabla_{\rvphi,\rvtheta}  \mathcal{L}(\rvx; \rvphi, \rvtheta)$
    \UNTIL{converged}
  \end{algorithmic}
\end{algorithm}
\end{minipage}
\hfill
\begin{minipage}[t]{0.47\textwidth}
\begin{algorithm}[H]
  \caption{Sampling} \label{alg:sampling}
  \small
  \begin{algorithmic}[1]
    \vspace{.04in}
    \STATE Prepare  \\
     $t_0:=0<t_1<\cdots<t_{N}:=1$ and \\ $\rvz_{t_N}\sim \mathcal{N}(\mathbf{0}, \mathbf{I})$
    \FOR {$n = N, N-1,\cdots,1$}
      \STATE $\rvz_{t_{n-1}} \sim p_{\rvphi, \rvtheta}(\rvz_{t_{n-1}}| \rvz_{t_{n}}; t_{n-1}, t_{n})$
    \ENDFOR
    \STATE  $\rvx \sim P_{\rvtheta}(\rvx|\rvz_0; 0)$
    \STATE \textbf{return} $\rvx$
    \vspace{.04in}
  \end{algorithmic}
\end{algorithm}
\end{minipage}
\vspace{-1em}
\end{figure*}

Algorithms~\ref{alg:training} and~\ref{alg:sampling} outline the training and sampling procedures of \method{}. For sampling, we discretize time $t\in[0,1]$ into $N+1$ points $\{t_n\}_{n=0}^N$ such that they satisfy $t_{n}<t_{n+1}$, $t_0=0$, and $t_N=1$. Starting with Gaussian noise sampled from $\rvz_{t_{N}}\sim \mathcal{N}(\bm{0}, \mathbf{I})$, we sample $\rvz_{0}$ through the ancestral sampling given by $p_{\rvphi, \rvtheta}(\rvz_{t_{n-1}}| \rvz_{t_{n}}; t_{n-1}, t_{n})$, which is defined in Eq.~\eqref{eq:reparameterization}. Finally, the discrete output $\rvx$ is obtained from the model $P_{\rvtheta}(\rvx|\rvz_{0}; 0)$. Note that, unlike CSDMs, our model directly outputs the token probabilities for continuous input $\rvz_{t_{n}}$ at timestep $t_{n}$.

\section{Additional Samples} \label{appendix:samples}
\begin{figure}[!htp]
    \centering
    \includegraphics[width=0.9\textwidth]{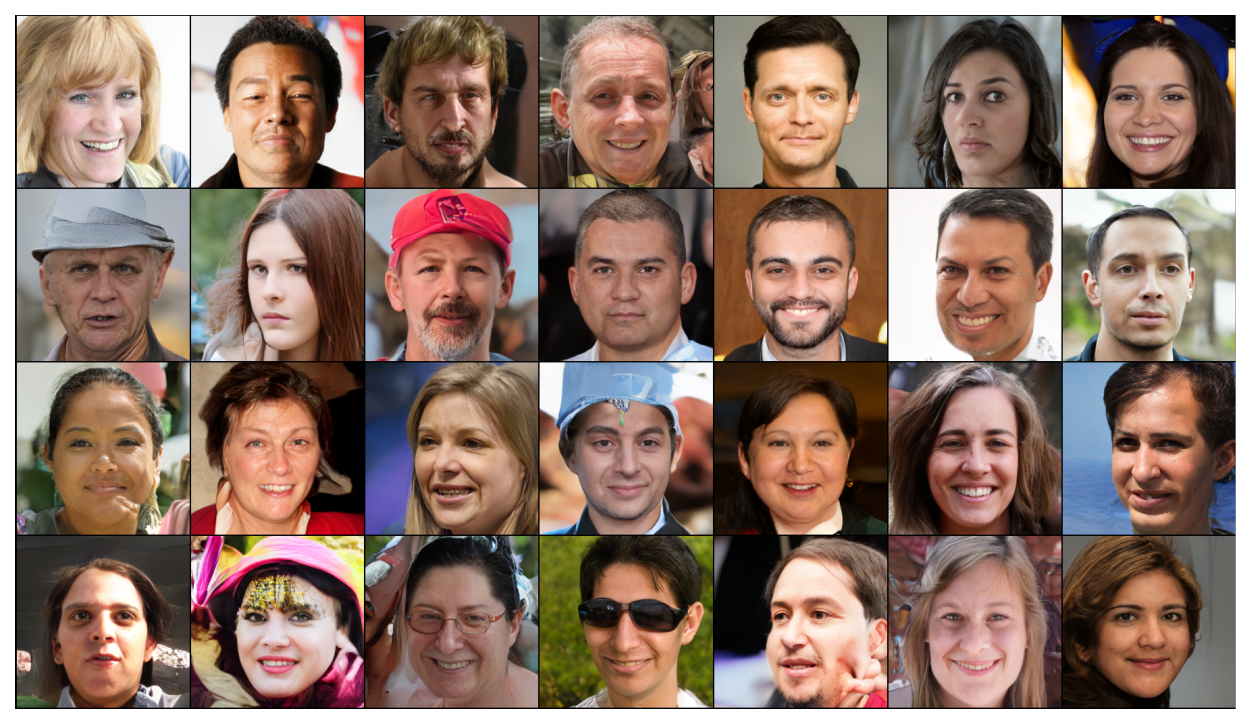}
    \caption{\method{} samples of unconditional image generation on FFHQ.}
    \label{fig:unconditional:ffhq}
\end{figure}

\begin{figure}[t]
    \centering
    \includegraphics[width=0.9\textwidth]{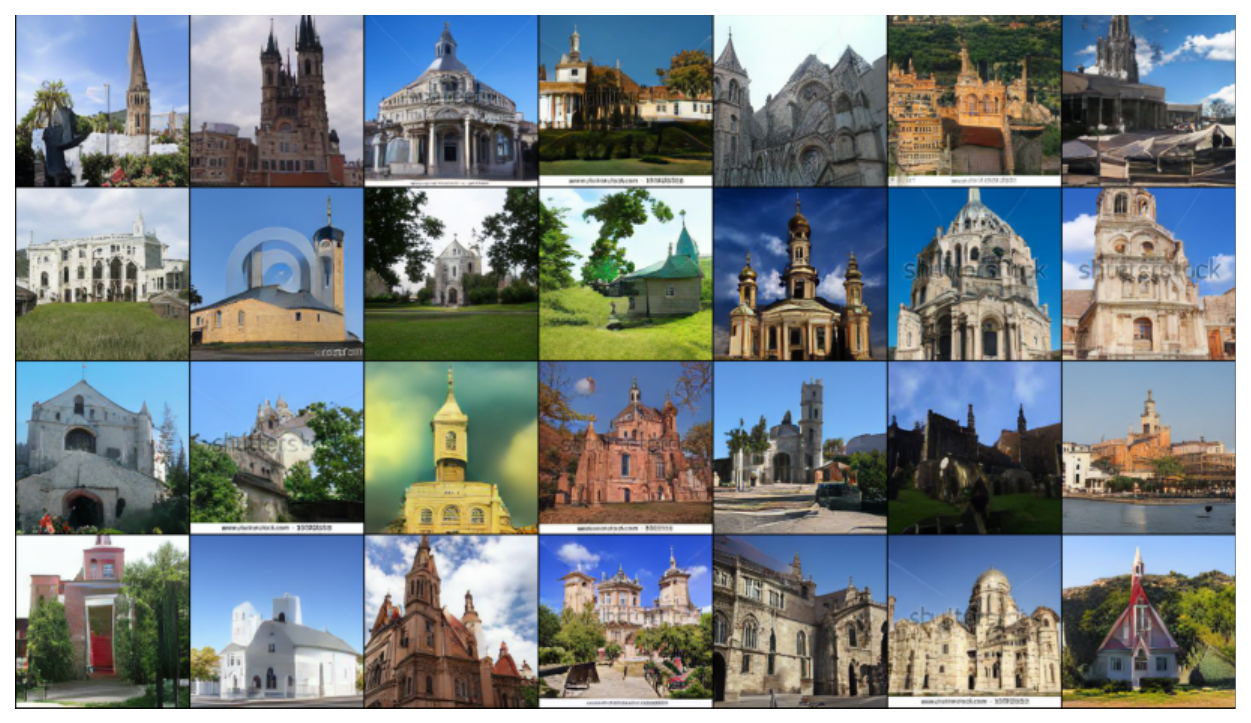}
    \caption{\method{} samples of unconditional image generation on LSUN Churches.}
    \label{fig:unconditional:churches}
\end{figure}

\begin{figure}[t]
    \centering
    \includegraphics[width=0.9\textwidth]{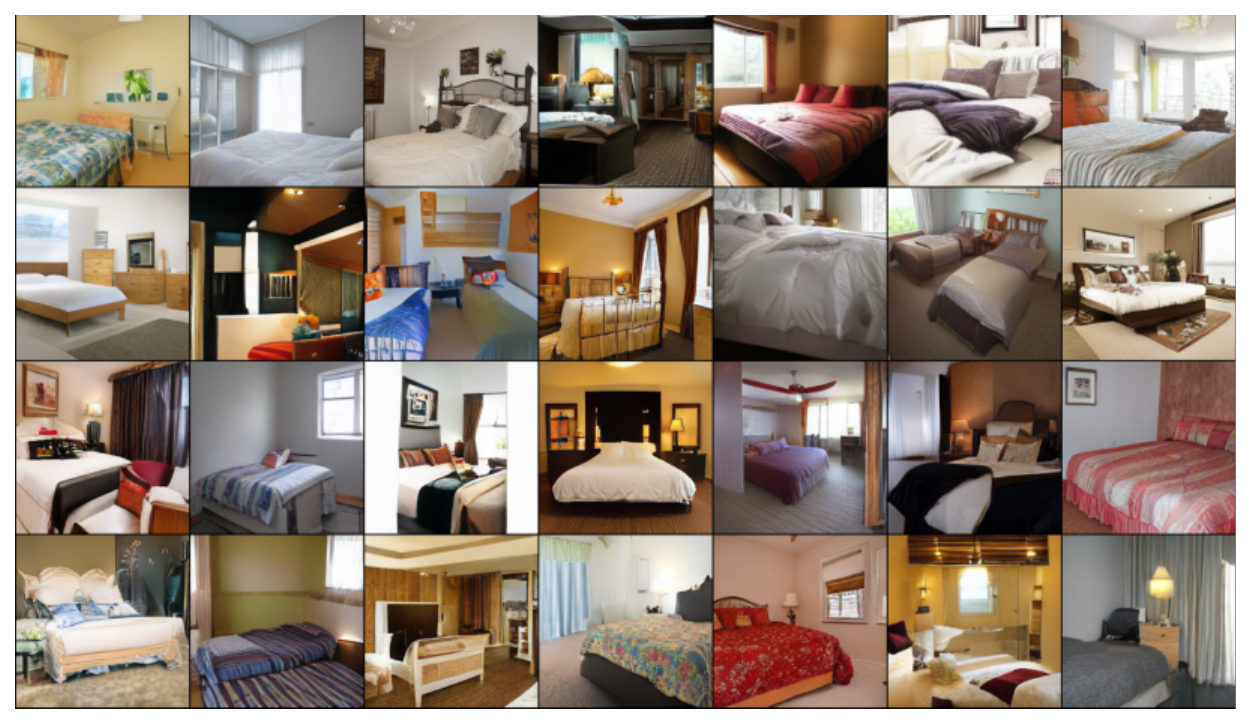}
    \caption{ \method{} samples of unconditional image generation on LSUN Bedrooms.}
    \label{fig:unconditional:bedroms}
\end{figure}

\begin{figure}[!htp]
    \centering
    \includegraphics[width=0.32\linewidth]{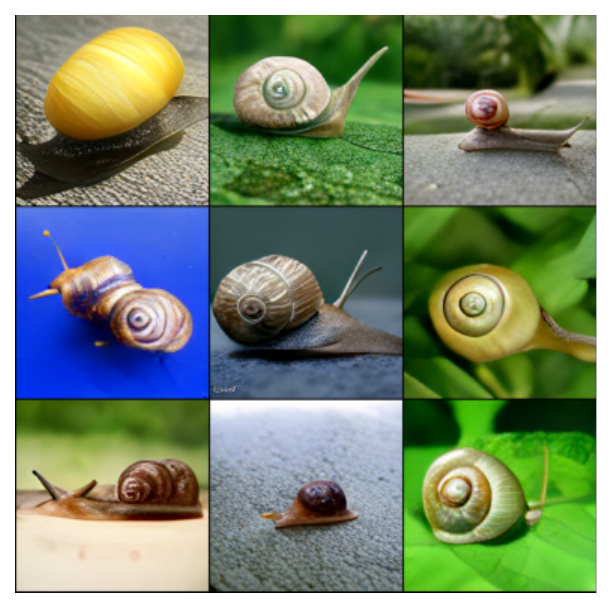}
    \includegraphics[width=0.32\linewidth]{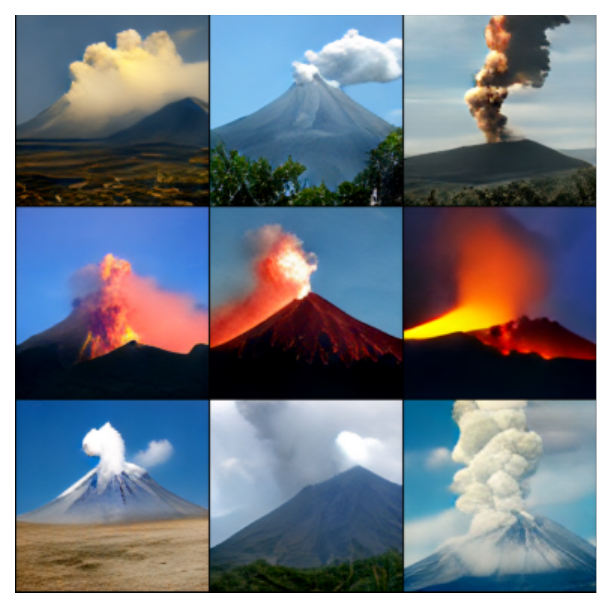}
    \includegraphics[width=0.32\linewidth]{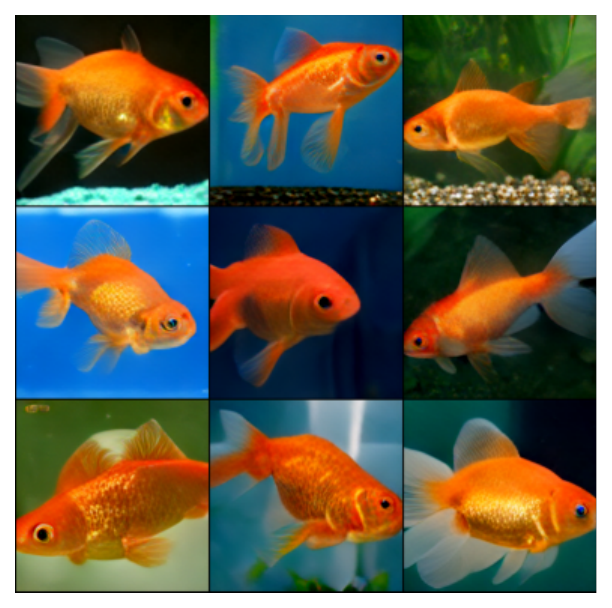} \\

    \includegraphics[width=0.32\linewidth]{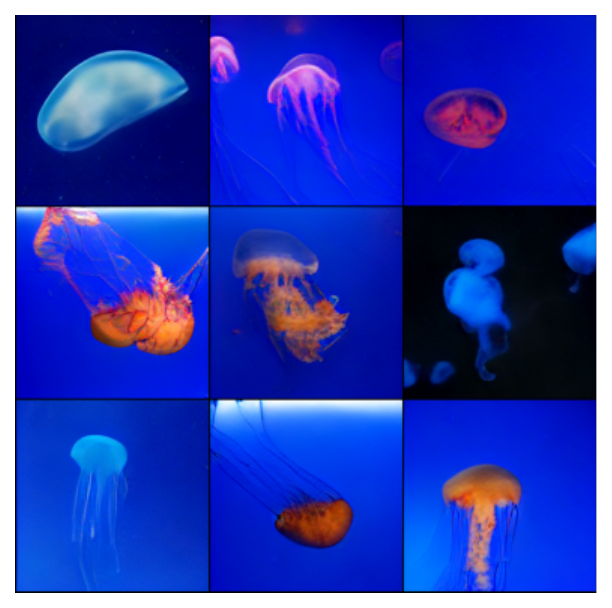}
    \includegraphics[width=0.32\linewidth]{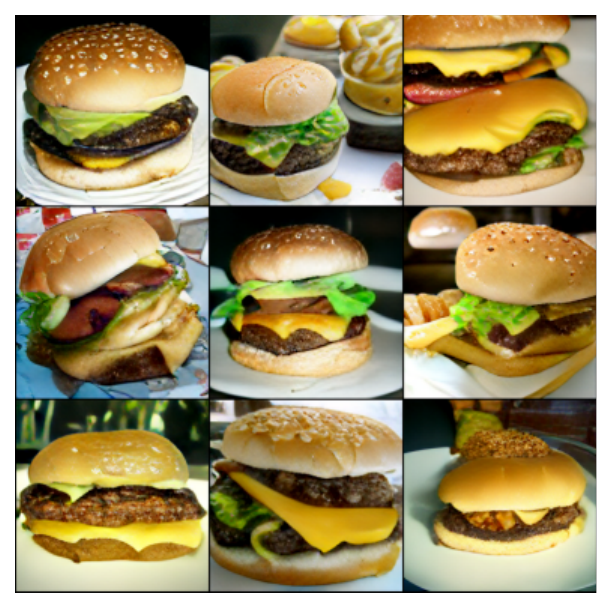}
    \includegraphics[width=0.32\linewidth]{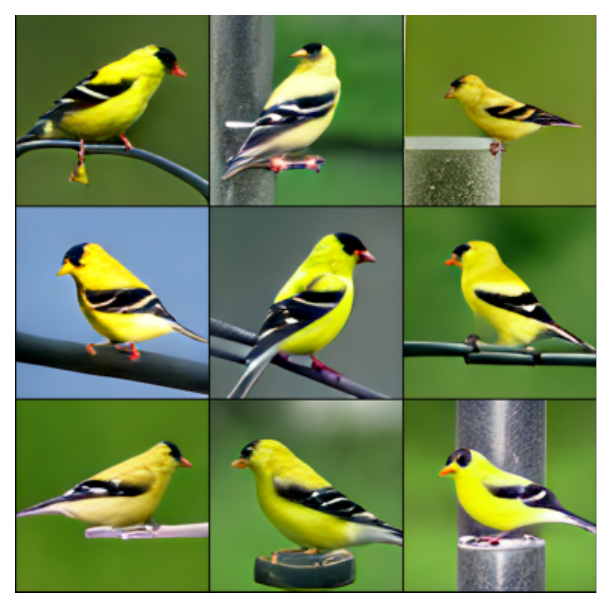}
    \caption{\method{} samples of conditional image generation on ImageNet $256\times 256$ for selected classes, including ``snail", ``volcano", ``goldfish", ``jellyfish", ``cheeseburger", ``goldfinch"}
    \label{fig:conditional:imagenet}
\end{figure}

In this section, we present additional samples generated by \method{}. For unconditional image generation, Figures~\ref{fig:unconditional:ffhq}, \ref{fig:unconditional:churches}, and \ref{fig:unconditional:bedroms} show the generated samples from \method{} trained on FFHQ, LSUN Churches, and LSUN Bedrooms, respectively. Figure~\ref{fig:conditional:imagenet} visualizes the conditional samples from ImageNet. All images are at a resolution of $256 \times 256$.


\end{document}

%% file: math_commands.tex
\usepackage{amsmath,amsfonts,bm}

\def\eqref#1{equation~\ref{#1}}

\def\1{\bm{1}}

\def\rvepsilon{{\mathbf{\epsilon}}}
\def\rvtheta{{\mathbf{\theta}}}

\def\rve{{\mathbf{e}}}

\def\rvm{{\mathbf{m}}}

\def\rvx{{\mathbf{x}}}
\def\rvy{{\mathbf{y}}}
\def\rvz{{\mathbf{z}}}

\def\va{{\bm{a}}}
\def\vb{{\bm{b}}}

\def\ve{{\bm{e}}}

\def\vh{{\bm{h}}}

\def\evx{{x}}

\def\mI{{\bm{I}}}

\DeclareMathAlphabet{\mathsfit}{\encodingdefault}{\sfdefault}{m}{sl}
\SetMathAlphabet{\mathsfit}{bold}{\encodingdefault}{\sfdefault}{bx}{n}

\newcommand{\E}{\mathbb{E}}

